\documentclass[authoryear,preprint,12pt]{elsarticle}

\usepackage{microtype}
\usepackage{subcaption}
\usepackage{booktabs}
\usepackage{tabularx}
\usepackage{float}
\usepackage{multirow}
\usepackage{mathtools}
\usepackage{amssymb}
\usepackage{amsthm}
\usepackage{makecell}

\usepackage[capitalize,noabbrev]{cleveref}

\theoremstyle{plain}

\theoremstyle{definition}

\theoremstyle{remark}

\journal{Neural Networks}

\begin{document}

\begin{frontmatter}

\title{Adaptive Cortically Constrained EEG--Vision Alignment for Zero-Shot Brain-to-Image Retrieval}


\author[aaa]{Ye Wang}
\author[aaa]{Haokun Ren}
\author[bbb]{Wei Wu}
\author[ccc]{Guoyin Wang\corref{cor1}}
\author[ddd]{Zhuliang Yu}
\author[aaa]{Hong Yu}
\author[aaa]{Ke Liu\corref{cor1}}

\cortext[cor1]{Corresponding authors: Guoyin Wang (wanggy@cqnu.edu.cn) and Ke Liu (liuke@cqupt.edu.cn) }

\affiliation[aaa]{organization={School of Artificial Intelligence, Chongqing University of Posts and Telecommunications},
    city={Chongqing},
    country={China}}

\affiliation[bbb]{organization={School of Medicine, Shanghai Jiaotong University},
    city={Shanghai},
    country={China}}

\affiliation[ccc]{organization={National Center for Applied Mathematics in Chongqing, Chongqing Normal University},
    city={Chongqing},
    country={China}}

\affiliation[ddd]{organization={School of Automation Science and Engineering, South China University of Technology},
    city={Guangzhou},
    country={China}}

\begin{abstract}
Zero-shot brain-to-image retrieval requires robust alignment between noisy EEG responses and visual representations. Existing EEG--vision alignment methods often operate in sensor space and apply fixed visual supervision to all responses, ignoring both spatial mixing in scalp EEG and response-wise variability in alignment reliability. We propose an adaptive cortically constrained EEG--vision alignment method for zero-shot brain-to-image retrieval. The method reconstructs EEG responses into predefined ROI-level source-pattern representations and encodes them with a Neuro-ROI Attention Encoder. To handle response-wise variability, we introduce an evidence-based adaptive visual supervision strategy that weights detail-controlled visual targets using model-based alignment evidence. On THINGS-EEG, the proposed method achieves strong 200-way zero-shot retrieval performance, with ROI-level attribution providing post hoc interpretability of the learned source-pattern representations. These results show that cortically constrained representation learning and adaptive supervision can jointly support EEG--vision alignment for zero-shot brain-to-image retrieval.
\end{abstract}


\end{frontmatter}

\section{Introduction}

Learning robust EEG--vision alignment is important for zero-shot brain-to-image retrieval, where a model must match an EEG response to the corresponding unseen image by comparing neural and visual representations \citep{palazzo2020decoding,du2023decoding,song2024decoding,li2024visual,wu2025bridging}. Electroencephalography (EEG) provides millisecond-level access to visual responses and offers a practical signal source for retrieval-oriented neural representation learning, but its low spatial resolution and response variability make cross-modal alignment challenging \citep{schirrmeister2017deep,lawhern2018eegnet,gifford2022large}. 

Most existing EEG--vision alignment approaches operate directly in sensor space, where scalp recordings reflect mixtures of activity from distributed cortical sources due to volume conduction \citep{song2024decoding,li2024visual,baillet2002electromagnetic}. This limits the use of anatomical structure as an inductive bias for representation learning and makes model attributions difficult to summarize beyond electrodes. At the same time, visual embeddings from modern pretrained models often contain fine-grained image information, whereas stimulus-specific EEG responses vary substantially in alignment reliability \citep{wu2025bridging}. Treating all EEG responses as if they should align with the same high-detail visual target may impose unreliable supervision and obscure response-wise variability in EEG--vision alignment.

\begin{figure}[htb]
  \centering
  \includegraphics[width=\columnwidth,keepaspectratio]{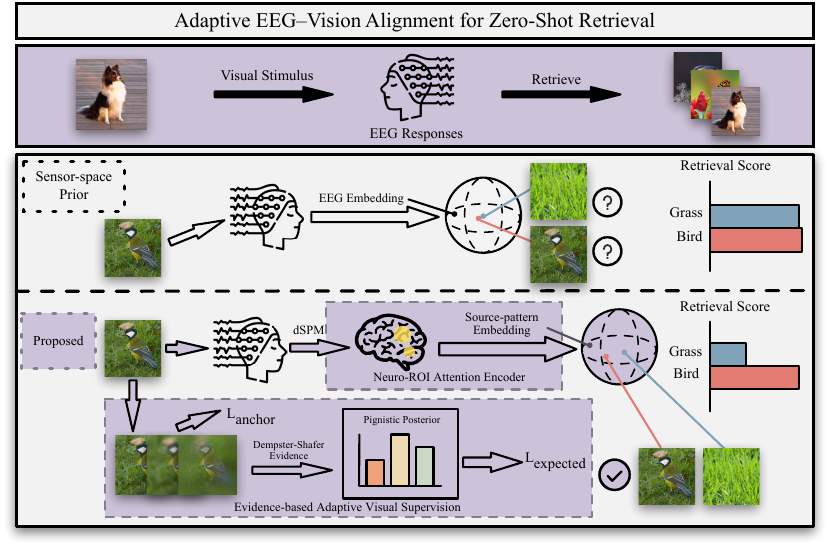}
  \caption{Motivation and overview of the proposed adaptive cortically constrained EEG--vision alignment method for zero-shot brain-to-image retrieval. Unlike conventional sensor-space EEG--vision alignment with fixed visual supervision, our method learns anatomically constrained ROI-level source-pattern representations and adaptively weights detail-controlled visual targets according to response-wise alignment evidence.}
  \label{fig:framework}
\end{figure}

These considerations motivate an adaptive EEG--vision alignment method that accounts for both cortical source constraints and response-wise variability in stimulus-evoked evidence. To this end, we propose a cortically constrained representation learning approach in which EEG responses are reconstructed into predefined cortical regions of interest (ROIs) and encoded with ROI-constrained attention. We then introduce an adaptive visual supervision strategy that weights visual targets with different levels of detail according to model-based response-wise alignment evidence, rather than assuming a fixed supervision granularity for all responses. In this way, responses with stronger alignment evidence can receive greater weight on high-detail targets, while more difficult-to-align responses receive more conservative supervision. Our contributions can be summarized below: 

(1) We propose a cortically constrained EEG source-pattern representation learning framework that transforms sensor-space EEG into ROI-level source-pattern representations for EEG--vision alignment;

(2) We introduce a response-wise adaptive visual supervision strategy that weights detail-controlled visual targets according to model-based alignment evidence;

(3) We demonstrate strong zero-shot brain-to-image retrieval performance on THINGS-EEG and provide ROI-level attribution as post hoc interpretability analysis of the learned source-pattern representations.

\section{Related Work}

\subsection{EEG--Vision Alignment and Brain-to-Image Retrieval}

EEG has been widely used as a non-invasive signal source for visual decoding because of its high temporal resolution. Recent advances in deep learning and pretrained vision models have enabled EEG-based object classification, image retrieval, and visual reconstruction \citep{schirrmeister2017deep,lawhern2018eegnet,spampinato2017deep,palazzo2020decoding,bai2023dreamdiffusion,guo2025neuro}. In particular, contrastive EEG--vision alignment methods learn a shared representation space between EEG responses and visual embeddings, providing an effective route for zero-shot brain-to-image retrieval \citep{radford2021learning,du2023decoding,song2024decoding}. However, most existing approaches operate directly on sensor-space EEG \citep{li2024visual,wei2024mb2c,chen2024visual,zhang2025cognitioncapturer}. Although sensor-level decoding can achieve promising performance, scalp recordings reflect mixtures of activity from distributed cortical sources due to volume conduction \citep{baillet2002electromagnetic}, making it difficult to impose anatomical structure on learned representations or summarize model attributions beyond electrodes.

\subsection{Source-Constrained EEG Representation Learning}

EEG source reconstruction provides a way to project scalp recordings into anatomically constrained cortical estimates and has been widely used to model putative cortical patterns of cognitive and perceptual processes \citep{hamalainen1994interpreting,dale2000dynamic,Wang2024Cortical,kang2026decoding,Schreiner2026Increasing}. Compared with sensor-space analysis, source-space modeling can mitigate, but not eliminate, ambiguity from volume conduction and allows EEG responses to be summarized within predefined cortical regions of interest \citep{baillet2002electromagnetic,glasser2016multi}. For neural-network-based EEG--vision alignment, such source constraints can serve as structured inductive biases and can support ROI-level interpretation of learned source-pattern representations. However, cortically constrained ROI representations have been less explored in recent EEG--vision alignment models, which are often optimized primarily for decoding performance using sensor-space inputs.

\subsection{Vision Encoders as Cross-Modal Supervision Targets}

Pretrained visual models, such as convolutional neural networks, vision transformers, and vision-language models, provide structured representations of natural images and have become useful computational tools for relating brain activity to visual information \citep{he2016deep,dosovitskiy2020image,caron2021emerging,oquab2023dinov2,radford2021learning}. Their feature spaces capture different levels of visual structure, ranging from local texture and shape information to higher-level object and semantic representations. In EEG--vision alignment, these visual embeddings are often used as supervision targets for neural representation learning \citep{du2023decoding,song2024decoding}. However, pretrained visual embeddings often contain fine-grained information that may not be equally reliable for every stimulus-specific EEG response \citep{wu2025bridging}. Directly aligning all EEG responses to a fixed high-detail visual representation may impose overly strong supervision, especially for difficult-to-align responses.

\subsection{Adaptive Supervision and Evidential Learning}

Stimulus-specific EEG responses vary substantially across responses and subjects because of neural variability, measurement noise, attention fluctuations, and differences in stimulus-evoked response strength \citep{schirrmeister2017deep,gifford2022large}. As a result, different responses may provide different degrees of alignment evidence for visual targets with different amounts of available image detail \citep{wu2025bridging}. Existing methods typically use a fixed visual target or apply the same supervision strategy to all responses, which ignores this variability \citep{du2023decoding,song2024decoding,li2024visual}. Recent work has begun to explore uncertainty-aware methods for EEG-based visual decoding, and evidential learning provides a useful framework for modeling uncertainty and ignorance under limited or conflicting evidence \citep{sensoy2018evidential,Shao2024Dual,Huang2025Deep,wu2025bridging}. However, the relationship between response-wise variability, visual-detail target selection, and cortically constrained ROI-level interpretability remains underexplored.

\begin{figure}[t!]
\centering
\includegraphics[width=\columnwidth,keepaspectratio]{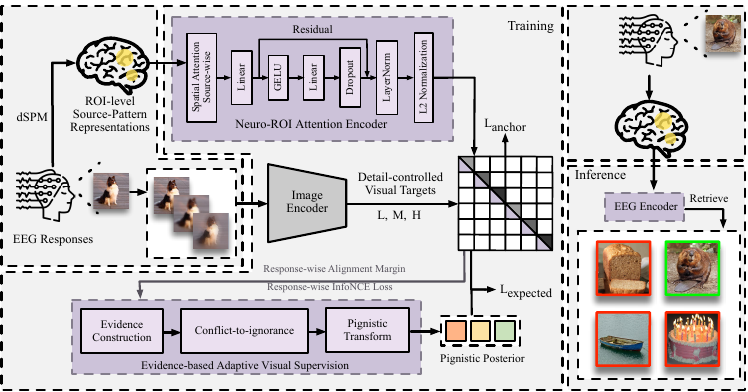}
\caption{Overview of the proposed adaptive cortically constrained EEG--vision alignment method. EEG responses are reconstructed into anatomically constrained ROI-level source-pattern representations and encoded by the Neuro-ROI Attention Encoder, which uses ROI-constrained source-wise attention to produce source-pattern embeddings. The evidence-based adaptive visual supervision strategy assigns response-wise weights to detail-controlled visual targets according to model-based alignment evidence, enabling adaptive EEG--vision alignment under variable response-wise evidence.}
\label{fig:method}
\end{figure}

\section{Method}

\subsection{Problem Definition}

Our task is to model EEG--vision alignment between EEG responses and visual embeddings in a zero-shot brain-to-image retrieval setting.
Let the training set be $\mathcal{D}_{\mathrm{tr}}=\{(\mathbf{I}_i,\mathbf{X}_i,u_i)\}_{i=1}^{N_{\mathrm{tr}}}$, where $\mathbf{I}_i$ is the stimulus image, $\mathbf{X}_i$ is the synchronized EEG response, and $u_i\in\mathcal{U}_{\mathrm{tr}}$ is the image category label.
The model is trained only on $\mathcal{D}_{\mathrm{tr}}$.
At test time, we evaluate on test set $\mathcal{D}_{\mathrm{te}}=\{(\mathbf{I}_j,\mathbf{X}_j,u_j)\}_{j=1}^{N_{\mathrm{te}}}$ with $u_j\in\mathcal{U}_{\mathrm{te}}$ and $\mathcal{U}_{\mathrm{tr}}\cap\mathcal{U}_{\mathrm{te}}=\emptyset$.
Given a test EEG sample $\mathbf{X}^\star$, the goal is to retrieve its corresponding image $\mathbf{I}^\star$ among candidates in $\mathcal{U}_{\mathrm{te}}$ by ranking cosine similarities between the EEG source-pattern embedding and image embeddings. Our overall framework is illustrated in \cref{fig:method}. Additional details are provided in \ref{sec:implementation}.

\subsection{Cortically Constrained ROI-level Source Representation}
Scalp EEG recordings are spatially smooth mixtures due to volume conduction, which combines activity from multiple putative cortical sources and reduces identifiability. To formulate EEG--vision alignment within an anatomically constrained ROI-level source representation, we perform source reconstruction and restrict the representation to source locations within predefined ROIs. 
For each sample, denote the raw (non-MVNN) EEG as $\mathbf{X}\in\mathbb{R}^{E\times T}$, where $E$ is the number of electrodes and $T$ the number of time points.
Under the linear forward model:
\begin{equation}
\mathbf{X} = \mathbf{G}\mathbf{s} + \boldsymbol{\epsilon},
\end{equation}

where $\mathbf{G}\in\mathbb{R}^{E\times S}$ is the lead field matrix, $\mathbf{s}\in\mathbb{R}^{S\times T}$ is the cortical source time series (under a fixed-orientation constraint, $S$ is the number of source locations), and $\boldsymbol{\epsilon}\in\mathbb{R}^{E\times T}$ is noise. 

We estimate the noise covariance $\mathbf{C}_n$ using only the pre-stimulus baseline segment. Then we estimate subject-wise covariances within each split, obtaining $\mathbf{C}_n^{\mathrm{tr}}$ and $\mathbf{C}_n^{\mathrm{te}}$, and construct the corresponding minimum-norm estimate (MNE) inverse operators $\mathbf{W}_{\mathrm{MNE}}^{\mathrm{tr}}$ and $\mathbf{W}_{\mathrm{MNE}}^{\mathrm{te}}$ \citep{hamalainen1994interpreting}. 
Because the covariance is estimated only from stimulus-free baseline segments, this procedure does not use evoked responses, visual targets, or category labels. Given the leadfield matrix $\mathbf{G}$ and regularization parameter $\lambda$, we compute the MNE inverse operator as:
\begin{equation}
\mathbf{W}_{\mathrm{MNE}}
=
\left(\mathbf{G}^\top \mathbf{C}_n^{-1}\mathbf{G}+\lambda \mathbf{I}^{(S)}\right)^{-1}
\mathbf{G}^\top \mathbf{C}_n^{-1},
\end{equation}

where $\mathbf{I}^{(S)}$ is the identity matrix of size $S$. Then we obtain the source estimate $\hat{\mathbf{s}}^{\mathrm{MNE}}_t=\mathbf{W}_{\mathrm{MNE}}\mathbf{x}_t$ at time $t$ from scalp measurements $\mathbf{x}_t$.
To reduce depth bias, we apply dSPM (dynamic statistical parametric mapping) normalization to the MNE solution \citep{dale2000dynamic}. Specifically, the standardized activity $\hat{s}^{\mathrm{dSPM}}_{j,t}$ at source $j$ and time $t$ is:
\begin{equation}
\hat{s}^{\mathrm{dSPM}}_{j,t}=
\frac{\hat{s}^{\mathrm{MNE}}_{j,t}}
{\sqrt{\left[\mathbf{W}_{\mathrm{MNE}}\mathbf{C}_n\mathbf{W}_{\mathrm{MNE}}^\top\right]_{jj}}}.
\end{equation}

We then select bilateral predefined ROIs according to the HCP-MMP1 parcellation and form the anatomically constrained ROI-level source estimates $\mathbf{Y}\in\mathbb{R}^{S'\times T}$, where $S'$ denotes the number of source locations within predefined ROIs \citep{glasser2016multi}. 
All subsequent encoding operates on this ROI-level source representation $\mathbf{Y}$, and ROI selection details are provided in \ref{sec:roi}. Because individual structural MRIs were not available, these source estimates are treated as ROI-pattern representations for downstream EEG--vision alignment and attribution analysis.

\subsection{Detail-Controlled Visual Targets as an Operational Probe}
\label{sec:mrvt}
To construct detail-controlled visual supervision targets, we generate a set of visual targets by applying controlled image blurring before extracting visual embeddings. This manipulation is used as an operational probe of response-wise alignment evidence, rather than as a direct measurement of neural visual-detail representation. Inspired by UBP~\citep{wu2025bridging}, we generate detail-controlled visual targets using a foveated blur transform~\citep{geisler1998real}. Given an image $\mathbf{I}$, for each detail-controlled target level 
$k\in\{L,M,H\}$, we first compute a uniformly blurred image 
$\mathbf{I}_{\mathrm{blur}}^{(k)}$ by convolving $\mathbf{I}$ 
with a disk kernel of radius $r_k$, where 
$r_H < r_M < r_L$. Here, $H$, $M$, and $L$ denote high-, medium-, 
and low-detail targets, respectively; larger kernel radii produce 
stronger blur and reduce the availability of fine visual details.

\begin{equation}
\mathbf{I}_{\mathrm{blur}}^{(k)}(i,j)
=
\sum_{u=-r_k}^{r_k}\sum_{v=-r_k}^{r_k}
\mathbf{I}(i-u,\,j-v)\cdot K_{\mathrm{disk}}^{(k)}(u,v),
\end{equation}

where $(i,j)$ denotes a pixel location, $(u,v)$ indexes kernel offsets, and $K_{\mathrm{disk}}^{(k)}(u,v)$ is the disk kernel, respectively:
\begin{equation}
\begin{gathered}
K_{\mathrm{disk}}^{(k)}(u,v)=
\begin{cases}
\frac{1}{Z_k}, & u^2+v^2\le r_k^2,\\
0, & \text{otherwise},
\end{cases}
\end{gathered}
\end{equation}

where $Z_k=\left|\{(u,v):u^2+v^2\le r_k^2\}\right|$ is the number of discrete offsets inside the radius-$r_k$ disk, so that $\sum_{u,v}K_{\mathrm{disk}}^{(k)}(u,v)=1$.

To mimic foveated vision, we blend the original image $\mathbf{I}$ with its blurred version $\mathbf{I}_{\mathrm{blur}}^{(k)}$ using a spatially varying mixing map $\alpha(i,j)=\exp\!\left(-\lambda\cdot\frac{\left\|(i,j)-(i_0,j_0)\right\|_2}{L}\right)$, yielding the foveated image $\tilde{\mathbf{I}}^{(k)}$:
\begin{equation}
\begin{gathered}
\tilde{\mathbf{I}}^{(k)}(i,j)
=
\alpha(i,j)\cdot \mathbf{I}(i,j)
+
\big(1-\alpha(i,j)\big)\cdot \mathbf{I}_{\mathrm{blur}}^{(k)}(i,j),
\end{gathered}
\end{equation}
where $(i_0,j_0)$ is the image center, $L$ is the maximum possible distance within the image, and $\lambda=0.3$ controls the radial decay rate of $\alpha$.

Finally, we extract a $d$-dimensional visual embedding from each foveated image using a frozen vision encoder $\phi$ (DINO-RN50):
\begin{equation}
\mathbf{t}^{(k)}=\phi\!\left(\tilde{\mathbf{I}}^{(k)}\right)\in\mathbb{R}^{d},\quad k\in\{L,M,H\}.
\end{equation}
The resulting target set $\{\mathbf{t}^{(L)},\mathbf{t}^{(M)},\mathbf{t}^{(H)}\}$ serves as detail-controlled visual targets for the evidence-based adaptive visual supervision strategy (Sec.~\ref{sec:elra}).

\subsection{Neuro-ROI Attention Encoder}
Given anatomically constrained ROI-level source estimates $\mathbf{Y}\in\mathbb{R}^{S'\times T}$, we introduce an ROI-constrained source-wise attention mechanism to further constrain the spatial hypothesis space and emphasize model-relevant source locations within predefined ROIs. Specifically, we learn a parameter vector $\boldsymbol{\alpha}\in\mathbb{R}^{S'}$ and convert it into a simplex-valued weight vector $\mathbf{w}\in\mathbb{R}^{S'}$ via softmax:
\begin{equation}
\mathbf{w}=\mathrm{softmax}(\boldsymbol{\alpha})\in\mathbb{R}^{S'}.
\end{equation}

Then we apply $\mathbf{w}$ to reweight source channels of $\mathbf{Y}$:
\begin{equation}
\tilde{\mathbf{Y}}=\mathrm{diag}(\mathbf{w})\,\mathbf{Y}\in\mathbb{R}^{S'\times T},
\end{equation}

where $\mathrm{diag}(\mathbf{w})\in\mathbb{R}^{S'\times S'}$ denotes the diagonal matrix whose diagonal entries are given by $\mathbf{w}$, respectively. 
Then we flatten $\tilde{\mathbf{Y}}$ across space and time,
$\mathbf{y}=\mathrm{vec}(\tilde{\mathbf{Y}})\in\mathbb{R}^{S'T}$,
and map it to a $d$-dimensional source-pattern embedding for EEG--vision alignment with a residual MLP projection:
\begin{equation}
\begin{gathered}
\mathbf{z} = \mathrm{LayerNorm}(\mathbf{u}) \in \mathbb{R}^{d},
\end{gathered}
\end{equation}

where $\mathbf{u} = \mathbf{h} + \mathrm{Dropout}\!\left(\mathbf{W}_1\,\mathrm{GELU}(\mathbf{h})+\mathbf{b}_1\right) \in \mathbb{R}^{d}$ and $\mathbf{h} = \mathbf{W}_0 \mathbf{y} + \mathbf{b}_0 \in \mathbb{R}^{d}$. The learned weights $\mathbf{w}$ quantify the relative contribution of anatomically constrained ROI-level source estimates, summarizing model-relevant spatial structure within the cortically constrained ROI-level source representation.

\subsection{Evidence-based Adaptive Visual Supervision Strategy}
\label{sec:elra}
In standard contrastive learning, each EEG source-pattern embedding is forced to align with a single visual target. However, alignment difficulty varies across responses: some responses provide weaker evidence for high-detail visual targets. We therefore use Dempster-Shafer theory (DST) to adaptively weight detail-controlled visual targets~\citep{dempster1968generalization,dempster2008upper,sensoy2018evidential}, which can explicitly represent ignorance under conflicting evidence. 

\textbf{DST mass function.} Let the frame of discernment be $\Theta=\{L,M,H\}$, corresponding to the three visual targets. We denote the ignorance set by $\Omega=\Theta$, and represent the mass function by its values on the singleton hypotheses $\{L\},\{M\},\{H\}$ and the full set $\Omega$. For each sample $i$, we maintain a basic probability assignment (BPA)
$\mathbf{m}_i=[m_i^{L},m_i^{M},m_i^{H},m_i^{\Omega}]$, where $m_i^{L},m_i^{M},m_i^{H}$ are masses assigned to $\{L\},\{M\},\{H\}$, and $m_i^{\Omega}$ is the mass assigned to $\Omega$ (ignorance):
\begin{equation}
\mathbf{m}_i = [m_i^{L}, m_i^{M}, m_i^{H}, m_i^{\Omega}],\quad
\sum_{A\in\{L,M,H,\Omega\}} m_i^{A}=1.
\end{equation}

\textbf{Evidence construction from training statistics.} DST specifies how to combine masses; the mapping from observable statistics to masses is a design choice. We construct two complementary sources of model-based alignment evidence from response-wise alignment statistics, using the medium-detail targets $\mathbf{t}^{(M)}$ as a stable reference. Let $\mathbf{z}_i\in\mathbb{R}^{d}$ denote the EEG source-pattern embedding for sample $i$ and $\mathbf{t}^{(M)}_j\in\mathbb{R}^{d}$ denote the medium-detail visual target for sample $j$ in the same batch. Let $\bar{\mathbf{z}}_i$ and $\bar{\mathbf{t}}^{(M)}_j$ be $\ell_2$-normalized embeddings. Define the temperature-scaled cosine similarity $s_{ij}=\frac{\bar{\mathbf{z}}_i^\top \bar{\mathbf{t}}^{(M)}_j}{\tau}$, where $\tau>0$ is a learnable temperature parameter.

We compute two scalar statistics for each response $i$:
(i) the per-response InfoNCE value $\ell_i=\mathrm{InfoNCE}\!\left(\bar{\mathbf{z}}_i,\bar{\mathbf{t}}^{(M)}_i;\{\bar{\mathbf{t}}^{(M)}_j\}_{j=1}^{B}\right)$, computed against the batch targets $\{\bar{\mathbf{t}}^{(M)}_j\}_{j=1}^{B}$, where larger $\ell_i$ indicates harder alignment and thus greater weight on lower-detail targets; and (ii) an alignment margin value $\mu_i= s_{ii} - \max_{j\neq i} s_{ij}$, where larger $\mu_i$ indicates easier separation and thus greater weight on higher-detail targets.

\textbf{Robust normalization.} Because the absolute scales of $\ell_i$ and $\mu_i$ may drift across subjects and training stages, we normalize each scalar $x\in\{\ell,\mu\}$ using EMA-smoothed running quantiles (30\%, 70\%) as robust scale references. Let $\theta_x^{(30)}$ and $\theta_x^{(70)}$ denote the EMA-smoothed running 30\% and 70\% quantiles of $\{x_i\}$, updated as $\theta_x^{(30)}\leftarrow \mathrm{EMA}\!\left(\mathrm{quantile}_{0.3}(\{x_i\})\right)$ and $\theta_x^{(70)}\leftarrow \mathrm{EMA}\!\left(\mathrm{quantile}_{0.7}(\{x_i\})\right)$, where $\mathrm{quantile}_q(\cdot)$ returns the $q$-th quantile, $\mathrm{EMA}(\cdot)$ denotes exponential moving average, and $\leftarrow$ indicates an update. We map each $x_i$ to $\tilde{x}_i\in[0,1]$:
\begin{equation}
\tilde{x}_i=
\mathrm{clamp}\!\left(
\frac{x_i-\theta_x^{(30)}}{\theta_x^{(70)}-\theta_x^{(30)}},
\,0,\,1
\right).
\end{equation}

\textbf{Statistic-to-evidence anchoring.}
We convert normalized statistics into soft detail-level evidence over the detail-controlled targets $\{L,M,H\}$ by mapping them to prototype distances $d$ on an ordinal detail scale, where $0$, $0.5$, $1$ correspond to $L$, $M$, $H$ respectively.
For the InfoNCE evidence, larger $\tilde{\ell}_i$ (harder sample) should favor low detail $L$:
\begin{equation}
\label{eq:dist_l_set}
\left\{d^{(\ell)}_{i,L},\, d^{(\ell)}_{i,M},\, d^{(\ell)}_{i,H}\right\}
=
\left\{1-\tilde{\ell}_i,\, \left|\tilde{\ell}_i-0.5\right|,\, \tilde{\ell}_i\right\}.
\end{equation}
For the margin evidence, larger $\tilde{\mu}_i$ (easier sample) should favor high detail $H$:
\begin{equation}
\label{eq:dist_mu_set}
\left\{d^{(\mu)}_{i,L},\, d^{(\mu)}_{i,M},\, d^{(\mu)}_{i,H}\right\}
=
\left\{\tilde{\mu}_i,\, \left|\tilde{\mu}_i-0.5\right|,\, 1-\tilde{\mu}_i\right\}.
\end{equation}
We map distances to categorical evidence $m^{(x)}_{i,k}$ via a temperature-controlled softmax:
\begin{equation}
m^{(x)}_{i,k}=\mathrm{softmax}_{k\in\{L,M,H\}}\!\left(-\frac{d^{(x)}_{i,k}}{T}\right), \quad x\in\{\ell,\mu\},
\end{equation}
where temperature $T=0.5$.

\textbf{Conflict-to-ignorance conversion.}
When the two sources of model-based alignment evidence conflict, we capture uncertainty via ignorance rather than making an overconfident decision. We measure their inconsistency $\mathrm{inc}_i$ by the half-$\ell_1$ discrepancy:
\begin{equation}
\mathrm{inc}_i = \frac{1}{2}\sum_{k\in\{L,M,H\}}\left|m_{i,k}^{(\ell)}-m_{i,k}^{(\mu)}\right|\in[0,1].
\end{equation}
We assign ignorance mass $m_i^{\Omega}$ as:
\begin{equation}
m_i^{\Omega}=\mathrm{clamp}\!\left(\beta_0+\beta_1\cdot \mathrm{inc}_i,\ \epsilon,\ 0.5\right),
\end{equation}
where $\beta_0=0.15$ and $\beta_1=0.3$ are fixed coefficients and $\epsilon=0.05$ prevents degenerate zeros, respectively. We cap $m_i^{\Omega}$ at $0.5$ to avoid degenerate cases where ignorance dominates and suppresses learning. The singleton masses $m_{i,k}$ are obtained by averaging the two evidence vectors and scaling by $(1-m_i^{\Omega})$:
\begin{equation}
m_{i,k}=\frac{m_{i,k}^{(\ell)}+m_{i,k}^{(\mu)}}{2}\cdot (1-m_i^{\Omega}),\quad
k\in\{L,M,H\}.
\end{equation}

\textbf{Sequential evidence fusion and discounting.}
To smooth predictions and accumulate evidence over training, we maintain a running historical BPA per sample and update it once per epoch by fusing it with the current BPA using Dempster's rule, assuming $m(\emptyset)=0$. We identify singleton sets with their labels (e.g., $L$ denotes $\{L\}$). Let $m_1$ and $m_2$ denote the historical and current mass functions, respectively. The conflict $K$ is:
\begin{equation}
K=\sum_{A\cap B=\emptyset} m_1(A)\,m_2(B),
\end{equation}
and the fused mass $m_{\mathrm{fused}}(C)$ is:
\begin{equation}
\begin{gathered}
m_{\mathrm{fused}}(C)=\frac{1}{1-K}\sum_{A\cap B=C} m_1(A)\,m_2(B),\\
C\in\{L,M,H,\Omega\}.
\end{gathered}
\end{equation}
After fusion, we apply epoch-wise discounting with $\delta$ to obtain the updated historical BPA for the next epoch: $m'(A)=\delta\,m(A),\ \forall A\neq\Omega$ and $m'(\Omega)=\delta\,m(\Omega)+(1-\delta)$, where $\delta=0.95$.

\textbf{Pignistic transform and expected contrastive loss.}
DST yields a mass function $\mathbf{m}_i$ in which $m_i^{\Omega}$ explicitly quantifies ignorance. Rather than making a hard decision that discards $m_i^{\Omega}$, we convert the BPA into a posterior over detail-controlled visual targets $q_{i,k}$ via the pignistic transform. The resulting posterior should be interpreted as a model-based response-wise alignment reliability estimate derived from alignment statistics, rather than as a direct ground-truth label of neural visual-detail representation:
\begin{equation}
q_{i,k}=m_i^{k}+\frac{m_i^{\Omega}}{|\Theta|},\quad k\in\{L,M,H\}.
\end{equation}

We then minimize the expected contrastive loss $\mathcal{L}_{\mathrm{expected}}$ over detail-controlled visual targets:
\begin{equation}
\mathcal{L}_{\mathrm{expected}}
=
\frac{1}{B}\sum_{i=1}^{B}\sum_{k\in\{L,M,H\}}
q_{i,k}\cdot \ell_{i}^{(k)},
\end{equation}

where $B$ is the batch size and $\ell_i^{(k)}$ is the InfoNCE loss computed between the normalized embeddings $\bar{\mathbf{z}}_i$ and $\bar{\mathbf{t}}^{(k)}_i$. In effect, responses with high ignorance mass yield flatter $q_{i,k}$ and thus more conservative supervision, while responses with stronger model-based response-wise alignment evidence concentrate posterior weight on the corresponding detail-controlled targets, supporting more stable alignment.

\subsection{Overall Objective}
Optimizing only the adaptive visual supervision objective can lead to a trivial solution where all responses collapse to the low-detail target, because the low-detail target reduces the availability of fine visual details and makes the alignment objective easier to minimize.
We therefore keep the medium-detail target as a stable anchor and enforce alignment on $M$ regardless of the response-wise posterior detail weights, where $\mathcal{L}_{\mathrm{SCE}}^{(M)}$ denotes the symmetric cross-entropy alignment loss computed at the medium-detail target $M$:
\begin{equation}
\mathcal{L}_{\mathrm{anchor}} = \mathcal{L}_{\mathrm{SCE}}^{(M)}.
\end{equation}

The final training loss is:
\begin{equation}
\mathcal{L}_{\mathrm{total}}
=
\mathcal{L}_{\mathrm{anchor}}
+
\lambda_{\mathrm{exp}}\cdot \mathcal{L}_{\mathrm{expected}}.
\end{equation}

We set $\lambda_{\mathrm{exp}}=0.2$ in all main experiments.

\section{Experiments and Results}

\subsection{Dataset and Implementation Details}
\textbf{THINGS-EEG} is a large-scale EEG dataset collected from 10 participants under a rapid serial visual presentation (RSVP) paradigm \citep{gifford2022large}. The training split contains 1,654 concepts with 10 images per concept; each image is presented 4 times per participant. The test split contains 200 concepts with one image per concept; each test image is presented 80 times per participant. Training and test concepts are disjoint. We largely follow the preprocessing pipeline of ATM-S \citep{li2024visual}, but exclude MVNN. MVNN applies covariance whitening across sensor channels, which alters sensor-space covariance and may be incompatible with the fixed lead field used for source reconstruction. Because we use cortically constrained ROI-level source modeling with anatomically predefined ROIs, we require source estimates that remain consistent with the lead field; therefore, we omit MVNN and instead apply dSPM-based source reconstruction followed by predefined cortical ROI selection. For these ROI-level source analyses, repetitions of the same stimulus were averaged after source reconstruction to improve the signal-to-noise ratio. Thus, each EEG response sample denotes a stimulus-specific, subject-specific averaged response representation rather than a raw presentation trial. Additional details are provided in \ref{sec:dataset}.

\textbf{Visual Encoders.} We use visual encoders pretrained with DINO and DINOv2 \citep{oquab2023dinov2}, covering multiple backbones including RN50, ViT-B/16, ViT-B/8, and ViT-B/14 \citep{caron2021emerging,dosovitskiy2020image}. Unless otherwise specified, all main-text experiments use a frozen DINO-RN50 image encoder. Effects of visual encoders are provided in \ref{sec:visual_encoders}.

\textbf{EEG Encoders.} To encode EEG responses, we use the proposed Neuro-ROI Attention Encoder. To more comprehensively evaluate the generalization of our approach, we also conduct experiments with several alternative EEG encoders, including ShallowNet \citep{schirrmeister2017deep}, DeepNet \citep{schirrmeister2017deep}, EEGNet \citep{lawhern2018eegnet}, TSConv \citep{song2024decoding}, and EEGProject \citep{wu2025bridging}. Comparisons of EEG encoders are provided in \ref{sec:brain_encoders}.

To ensure the stability of the experimental results, we conducted five independent training runs and report the average.

\subsection{Zero-shot Retrieval Comparisons as the Primary Evaluation Task}
We evaluate zero-shot brain-to-image retrieval as the primary downstream task for assessing EEG--vision alignment performance. We compare the proposed adaptive cortically constrained EEG--vision alignment method against recent EEG--vision alignment methods, including BraVL \citep{du2023decoding}, NICE \citep{song2024decoding}, ATM-S \citep{li2024visual}, CognitionCapturer (CogCap) \citep{zhang2025cognitioncapturer}, VE-SDN \citep{chen2024visual}, and UBP \citep{wu2025bridging}. Table~\ref{tab:table1} reports Top-1 and Top-5 accuracy for 200-way zero-shot brain-to-image retrieval on THINGS-EEG under both intra-subject and inter-subject settings. The proposed approach achieves 56.7\% Top-1 and 83.0\% Top-5 accuracy in the intra-subject setting, compared with 50.9\%/79.7\% for the strongest baseline UBP. In the inter-subject setting, our approach remains competitive in Top-1 accuracy but does not consistently improve Top-5 accuracy over all baselines. This pattern indicates competitive retrieval performance while additionally providing a post hoc ROI-level attribution route that conventional sensor-space EEG--vision alignment methods do not directly offer; it also suggests that cross-subject transfer remains limited under template-based source reconstruction.




\providecommand{\best}[1]{\textbf{#1}}

\begin{table}[htbp]
\caption{Top-1 and Top-5 accuracy (\%) of 200-way zero-shot brain-to-image retrieval on THINGS-EEG.}
\centering
\scriptsize
\setlength{\tabcolsep}{2.2pt}
\renewcommand{\arraystretch}{1.15}

\resizebox{\textwidth}{!}{%
\begin{tabular}{l*{22}{c}}
\toprule
& \multicolumn{2}{c}{S01}
& \multicolumn{2}{c}{S02}
& \multicolumn{2}{c}{S03}
& \multicolumn{2}{c}{S04}
& \multicolumn{2}{c}{S05}
& \multicolumn{2}{c}{S06}
& \multicolumn{2}{c}{S07}
& \multicolumn{2}{c}{S08}
& \multicolumn{2}{c}{S09}
& \multicolumn{2}{c}{S10}
& \multicolumn{2}{c}{Avg} \\
\cmidrule(lr){2-3}\cmidrule(lr){4-5}\cmidrule(lr){6-7}\cmidrule(lr){8-9}\cmidrule(lr){10-11}
\cmidrule(lr){12-13}\cmidrule(lr){14-15}\cmidrule(lr){16-17}\cmidrule(lr){18-19}\cmidrule(lr){20-21}\cmidrule(lr){22-23}
Method
& Top-1 & Top-5 & Top-1 & Top-5 & Top-1 & Top-5 & Top-1 & Top-5 & Top-1 & Top-5
& Top-1 & Top-5 & Top-1 & Top-5 & Top-1 & Top-5 & Top-1 & Top-5 & Top-1 & Top-5
& Top-1 & Top-5 \\
\midrule

\multicolumn{23}{c}{\textbf{Intra-Subject: train and test on one subject}}\\
\midrule
BraVL    &  6.1 & 17.9 &  4.9 & 14.9 &  5.6 & 17.4 &  5.0 & 15.1 &  4.0 & 13.4 &  6.0 & 18.2 &  6.5 & 20.4 &  8.8 & 23.7 &  4.3 & 14.0 &  7.0 & 19.7 &  5.8 & 17.5 \\
NICE     & 13.2 & 39.5 & 13.5 & 40.3 & 14.5 & 42.7 & 20.6 & 52.7 & 10.1 & 31.5 & 16.5 & 44.0 & 17.0 & 42.1 & 22.9 & 56.1 & 15.4 & 41.6 & 17.4 & 45.8 & 16.1 & 43.6 \\
NICE-SA  & 13.3 & 40.2 & 12.1 & 36.1 & 15.3 & 39.6 & 15.9 & 49.0 &  9.8 & 34.4 & 14.2 & 42.4 & 17.9 & 43.6 & 18.2 & 50.2 & 14.4 & 38.7 & 16.0 & 42.8 & 14.7 & 41.7 \\
NICE-GA  & 15.2 & 40.1 & 13.9 & 40.1 & 14.7 & 42.7 & 17.6 & 48.9 &  9.0 & 29.7 & 16.4 & 44.4 & 14.9 & 43.1 & 20.3 & 52.1 & 14.1 & 39.7 & 19.6 & 46.7 & 15.6 & 42.8 \\
ATM-S    & 25.6 & 60.4 & 22.0 & 54.5 & 25.0 & 62.4 & 31.4 & 60.9 & 12.9 & 43.0 & 21.3 & 51.1 & 30.5 & 61.5 & 38.8 & 72.0 & 34.4 & 51.5 & 29.1 & 63.5 & 28.5 & 60.4 \\
CogCap   & 27.2 & 59.5 & 28.7 & 57.0 & 37.2 & 66.1 & 37.7 & 63.2 & 21.8 & 47.8 & 31.6 & 58.1 & 32.8 & 59.6 & 47.6 & 73.5 & 33.4 & 57.7 & 35.1 & 63.6 & 33.3 & 60.6 \\
VE-SDN   & 32.6 & 63.7 & 34.4 & 69.9 & 38.7 & 73.5 & 39.8 & 72.0 & 29.4 & 58.6 & 34.5 & 68.8 & 34.5 & 68.3 & 49.3 & 79.8 & 39.0 & 69.6 & 39.8 & 75.3 & 37.2 & 69.9 \\
UBP      & 41.2 & 70.5 & 51.2 & \best{80.9} & 51.2 & 82.0 & 51.1 & 76.9 & \best{42.2} & \best{72.8} & 57.5 & 83.5 & 49.0 & 79.9 & 58.6 & 85.8 & 45.1 & 76.2 & 61.5 & 88.2 & 50.9 & 79.7 \\
\textbf{Ours}
         & \best{58.2} & \best{87.3} & \best{54.7} & 80.6 & \best{58.4} & \best{83.9} & \best{57.6} & \best{82.3} & 38.1 & 67.7
         & \best{63.3} & \best{83.6} & \best{56.3} & \best{83.7} & \best{59.4} & \best{86.0} & \best{55.1} & \best{85.0} & \best{65.4} & \best{90.1} & \best{56.7} & \best{83.0} \\
\midrule

\multicolumn{23}{c}{\textbf{Inter-Subject: leave one subject out for test}}\\
\midrule
BraVL    &  2.3 &  8.0 &  1.5 &  6.3 &  1.4 &  5.9 &  1.7 &  6.7 &  1.5 &  5.6 &  1.8 &  7.2 &  2.1 &  8.1 &  2.2 &  7.6 &  1.6 &  6.4 &  2.3 &  8.5 &  1.8 &  7.0 \\
NICE     &  7.6 & 22.8 &  5.9 & 20.5 &  6.0 & 22.3 &  6.3 & 20.7 &  4.4 & 18.3 &  5.6 & 22.2 &  5.6 & 19.7 &  6.3 & 22.0 &  5.7 & 17.6 &  8.4 & 28.3 &  6.2 & 21.4 \\
NICE-SA  &  7.0 & 22.6 &  6.6 & 23.2 &  7.5 & 23.7 &  5.4 & 21.4 &  6.4 & 22.2 &  7.5 & 22.5 &  3.8 & 19.1 &  8.5 & 24.4 &  7.4 & 22.3 &  9.8 & 29.6 &  7.0 & 23.1 \\
NICE-GA  &  5.9 & 21.4 &  6.4 & 22.7 &  5.5 & 20.1 &  6.1 & 21.0 &  4.7 & 19.5 &  6.2 & 22.5 &  5.9 & 19.1 &  7.3 & 25.3 &  4.8 & 18.3 &  6.2 & 26.3 &  5.9 & 21.6 \\
ATM-S    & 10.5 & 26.8 &  7.1 & 24.8 & \best{11.9} & \best{33.8} & 14.7 & \best{39.4} &  7.0 & 23.9 & 11.1 & \best{35.8}
         & \best{16.1} & \best{43.5} & \best{15.0} & \best{40.3} &  4.9 & 22.7 & \best{20.5} & 46.5 & 11.8 & \best{33.7} \\
UBP      & 11.5 & 29.7 & 15.5 & 40.0 &  9.8 & 27.0 & 13.0 & 32.3 & \best{8.8} & \best{33.8} & 11.7 & 31.0
         & 10.2 & 23.8 & 12.2 & 32.2 & \best{15.5} & \best{40.5} & 16.0 & 43.5 & 12.4 & 33.4 \\
\textbf{Ours}
         & \best{15.6} & \best{39.8} & \best{16.4} & \best{40.7} &  9.0 & 14.0 & \best{15.0} & 37.2 & 8.2 & 21.4
         & \best{13.1} & 31.8 & 10.3 & 29.0 & 13.4 & 31.1 & 8.2 & 22.2 & 20.3 & \best{47.8} & \best{13.0} & 31.5 \\
\bottomrule
\end{tabular}%
}
\label{tab:table1}
\end{table}

\subsection{Ablation Study on Evidence-based Adaptive Visual Supervision}
To diagnose which design choices contribute most, we perform a component-wise ablation by removing each module of the evidence-based adaptive visual supervision strategy individually while keeping all other components and training settings unchanged.
Specifically, we ablate: CIC (conflict-to-ignorance conversion, i.e., assigning ignorance mass $m_i^{\Omega}$ under conflicting evidence),
PT (pignistic transform that converts the BPA into $q_{i,k}$),
HFD (sequential historical evidence fusion with epoch-wise discounting),
and ECL (the expected contrastive loss $\mathcal{L}_{\mathrm{expected}}$).
As shown in Table~\ref{tab:ablation_remove}, the full evidence-based adaptive visual supervision strategy achieves the best performance. Removing individual modules leads to modest but consistent performance decreases, suggesting that these design choices jointly contribute to the EEG--vision alignment performance. Given the small absolute differences, we interpret the ablation primarily as evidence for the stability of the evidence-based adaptive visual supervision design rather than as proof that each component independently yields a large gain.


\begin{table}[htbp]
\caption{Component-wise ablation of the evidence-based adaptive visual supervision strategy on THINGS-EEG 200-way zero-shot retrieval by removing each module individually. Dashes indicate modules that are not applicable when the expected contrastive loss branch is removed.}
\centering
\small
\setlength{\tabcolsep}{6pt}
\begin{tabular}{cccccc}
\toprule
\multicolumn{4}{c}{\textbf{Evidence-based Adaptive Visual Supervision}} & \multicolumn{2}{c}{\textbf{Avg. Acc. (\%)}} \\
\cmidrule(lr){1-4}\cmidrule(lr){5-6}
\textbf{CIC} & \textbf{HFD} & \textbf{PT} & \textbf{ECL} & \textbf{Top-1} & \textbf{Top-5} \\
\midrule
$\times$     & $\checkmark$ & $\checkmark$ & $\checkmark$ & 56.3 $\pm$ 0.2 & 82.1 $\pm$ 0.3 \\
$\checkmark$ & $\times$     & $\checkmark$ & $\checkmark$ & 56.4 $\pm$ 0.1 & 82.5 $\pm$ 0.2 \\
$\checkmark$ & $\checkmark$ & $\times$     & $\checkmark$ & 56.4 $\pm$ 0.2 & 82.6 $\pm$ 0.2 \\
\textemdash  & \textemdash  & \textemdash  & $\times$     & 55.9 $\pm$ 0.3 & 82.7 $\pm$ 0.1 \\
$\checkmark$ & $\checkmark$ & $\checkmark$ & $\checkmark$ & \textbf{56.7} $\pm$ 0.2 & \textbf{83.0} $\pm$ 0.2 \\
\bottomrule
\end{tabular}
\label{tab:ablation_remove}
\end{table}

\subsection{Response-wise Posterior Stratification of Alignment Reliability}

To examine whether the evidence-based adaptive visual supervision strategy (Sec.~\ref{sec:elra}) provides meaningful model-based response-wise alignment evidence, we analyze the learned response-wise pignistic posterior $q_{i,k}$ and ignorance mass $m_i^{\Omega}$ across all 2,000 intra-subject test EEG responses (200 concepts $\times$ 10 subjects). This posterior analysis on the test set is conducted post hoc using the known EEG--image correspondence for interpretive analysis only; the posterior is not used during retrieval inference. Because the response-wise alignment posterior is derived from alignment statistics, its association with retrieval performance should be interpreted as a diagnostic and calibration analysis rather than as independent evidence for ground-truth neural visual-detail states.

\begin{figure}[!t]
\centering
\includegraphics[width=0.95\linewidth]{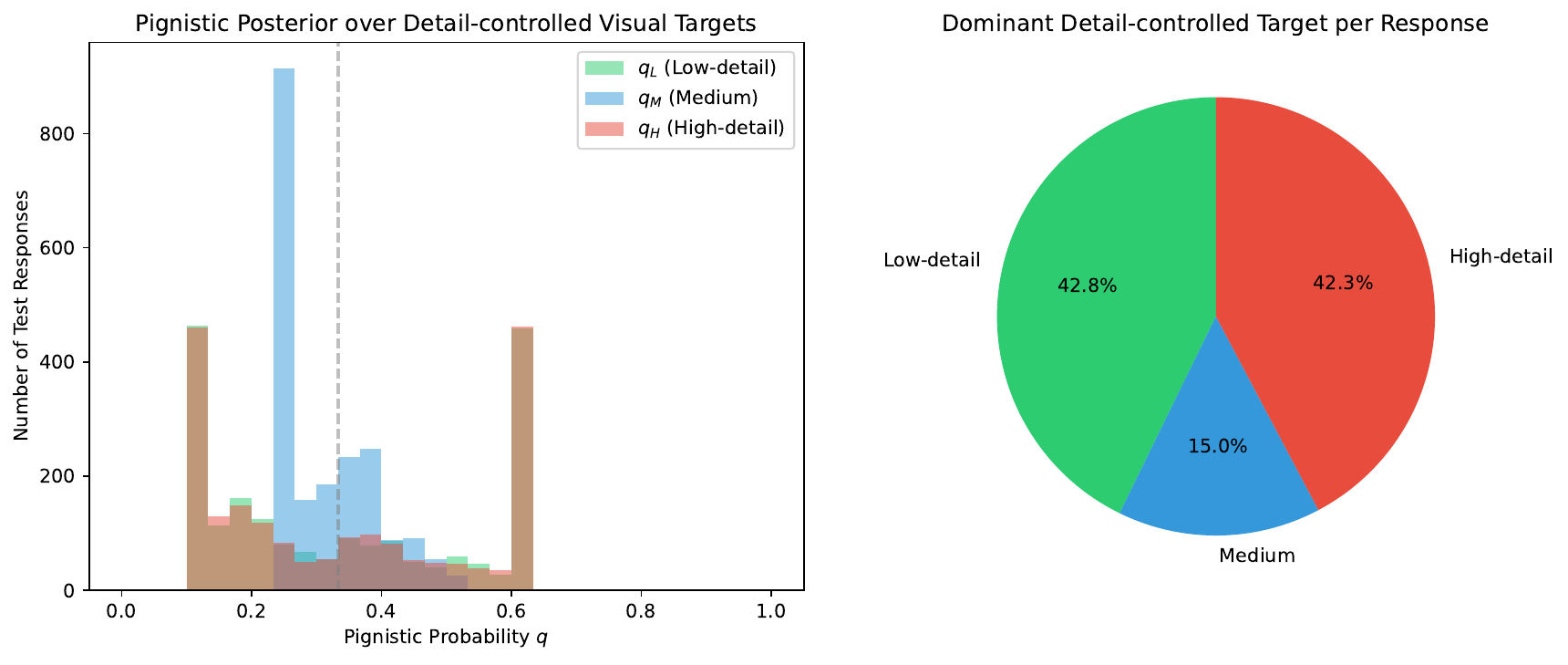}
\caption{Response-wise pignistic posterior distribution over detail-controlled visual targets. The left panel shows overlaid histograms of $q_L$, $q_M$, and $q_H$ across all responses, with the dashed vertical line marking the uniform prior ($q_k = 1/3$). The right panel shows the proportion of responses in which each detail-controlled visual target dominates.}
\label{fig:trial_q_dist}
\end{figure}

\textbf{Posterior Distribution over Detail-Controlled Visual Targets.}
As shown in \cref{fig:trial_q_dist}, across all test responses the mean posterior is $q_L = 0.341$, $q_M = 0.317$, $q_H = 0.342$, with an average response-wise standard deviation of 0.152. While the marginal means are close to the uniform prior ($q_k \approx 1/3$), the response-wise posteriors are strongly differentiated: in 42.8\% of responses $q_L$ dominates (low-detail visual target), in 42.3\% $q_H$ dominates (high-detail visual target), and only 15.0\% have $q_M$ as the winner. This pattern--marginal uniformity paired with sharp response-wise splits--is consistent with a model-based alignment-evidence tracker that assigns different responses to different regions of the detail-controlled visual-target space. It also partly reflects the ordinal evidence construction, in which the loss- and margin-based statistics tend to provide stronger evidence for the two endpoints under easy or difficult alignment conditions, while the medium target mainly serves as a stable anchor in the training objective.

\begin{figure}[!t]
\centering
\includegraphics[width=0.95\linewidth]{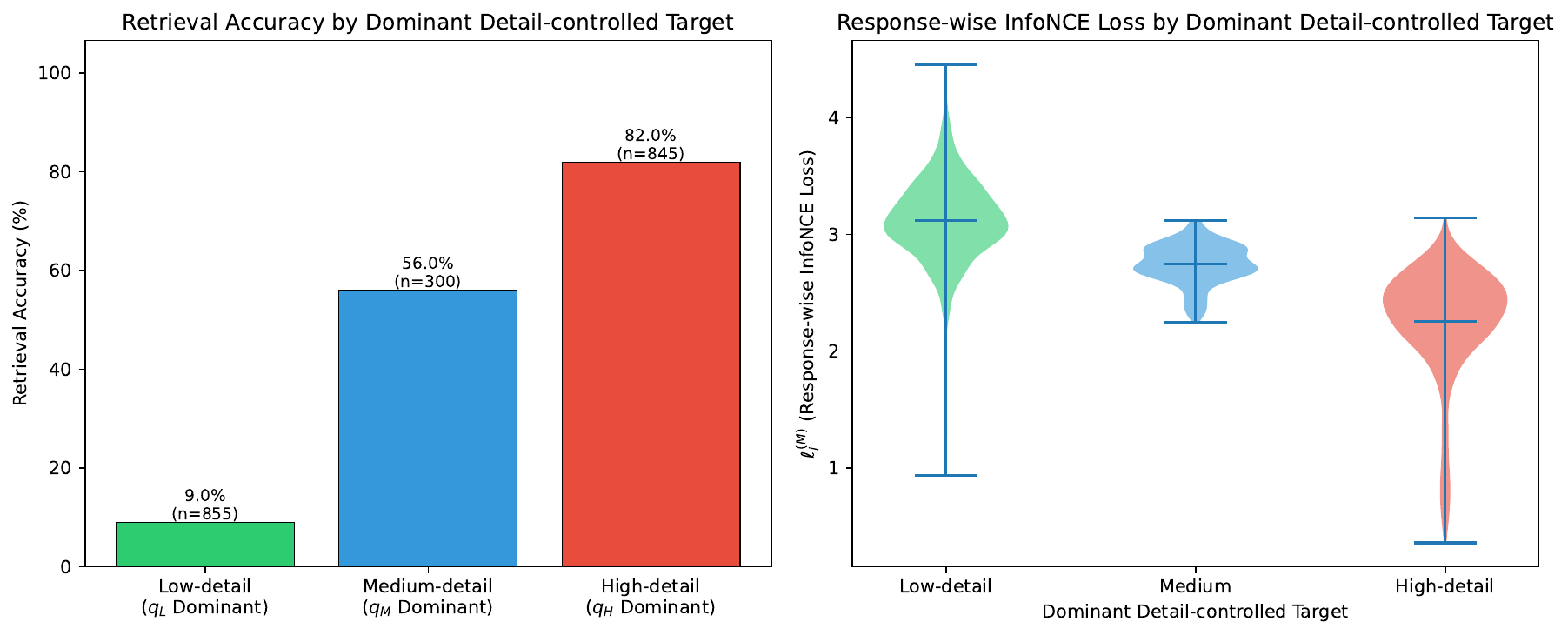}
\caption{Retrieval performance by dominant detail-controlled target. The left panel shows retrieval accuracy stratified by the dominant $q_k$ of each response, with $n$ values annotated above the bars. The right panel shows the distribution of response-wise medium-detail InfoNCE loss $\ell_i^{(M)}$ by dominant detail-controlled target.}
\label{fig:trial_retrieval_by_detail}
\end{figure}

\textbf{Retrieval accuracy by dominant detail-controlled target.}
\cref{fig:trial_retrieval_by_detail} reports retrieval accuracy stratified by each response's dominant $q_k$.
Responses dominated by $q_H$ (stronger model-based response-wise alignment evidence for the high-detail visual target) achieve \textbf{82.0\%} retrieval accuracy, those dominated by $q_M$ achieve \textbf{56.0\%}, and those dominated by $q_L$ (greater posterior weight on the low-detail visual target) achieve \textbf{9.0\%}.
The percentage-point gap between high-detail-dominant and low-detail-dominant responses indicates that the response-wise posterior is strongly associated with retrieval success and response-wise alignment difficulty.
The right panel further shows that response-wise InfoNCE loss increases monotonically from high-detail-dominant ($\bar{\ell}^{(M)}{=}2.25$) to low-detail-dominant responses ($\bar{\ell}^{(M)}{=}3.12$).

\textbf{Response-wise reliability stratification.}
We further partition test responses into three posterior-defined tiers: high-detail-evidence responses where $q_H > 0.5$ (stronger model-based response-wise alignment evidence for the high-detail visual target), low-detail/high-ignorance responses where $q_L > 0.4$ or $m_i^{\Omega} > 0.3$, and an intermediate group containing the remaining responses.
As reported in \cref{tab:trial_reliability}, high-detail-evidence responses ($n{=}582$, 29.1\%) achieve \textbf{86.9\%} retrieval accuracy with the lowest response-wise InfoNCE loss ($\bar{\ell}^{(M)}{=}2.14$) and lowest ignorance mass ($\bar{m}^{\Omega}{=}0.16$).
In contrast, low-detail/high-ignorance responses ($n{=}861$, 43.1\%) achieve \textbf{10.7\%} accuracy with substantially higher loss ($\bar{\ell}^{(M)}{=}3.10$). This tier mainly reflects greater evidence for low-detail alignment, with a subset also showing high ignorance.
The monotonic accuracy gradient across tiers (86.9\% $\to$ 61.0\% $\to$ 10.7\%) suggests that the response-wise posterior stratifies responses along a meaningful axis of model-based EEG--vision alignment reliability.


\begin{table}[htbp]
\caption{Response-wise stratification based on the evidence-based adaptive visual supervision posterior. High-detail evidence: $q_H > 0.5$ (stronger model-based response-wise alignment evidence for the high-detail visual target). Low-detail/high-ignorance: $q_L > 0.4$ or $m_i^{\Omega} > 0.3$ (greater low-detail evidence or high ignorance). Intermediate: remaining responses. $\bar{\ell}^{(M)}$ denotes mean response-wise InfoNCE loss at the medium-detail target; $\bar{m}^{\Omega}$ denotes mean ignorance mass; $\bar{q}_L$, $\bar{q}_M$, $\bar{q}_H$ denote mean pignistic posterior weights over low-, medium-, and high-detail visual targets.}
\centering
\footnotesize
\setlength{\tabcolsep}{4pt}
\begin{tabular}{lccrrrrr}
\toprule
\textbf{Tier} & \textbf{Responses} & \textbf{Acc. (\%)} & \textbf{$\bar{\ell}^{(M)}$} & \textbf{$\bar{m}^{\Omega}$} & \textbf{$\bar{q}_L$} & \textbf{$\bar{q}_M$} & \textbf{$\bar{q}_H$} \\
\midrule
High-detail evidence & 582 & 86.9 & 2.14 & 0.16 & 0.13 & 0.27 & 0.60 \\
Intermediate & 557 & 61.0 & 2.66 & 0.23 & 0.25 & 0.41 & 0.34 \\
\makecell[l]{Low-detail/\\high-ignorance} & 861 & 10.7 & 3.10 & 0.19 & 0.54 & 0.29 & 0.17 \\
\bottomrule
\end{tabular}
\label{tab:trial_reliability}
\end{table}

These results indicate that the evidence-based adaptive visual supervision strategy yields posteriors that are strongly associated with retrieval success and response-wise alignment difficulty. We therefore interpret the posterior as a model-based response-wise reliability measure for detail-controlled EEG--vision alignment, rather than as a direct categorical label of neural visual-detail representation.

\subsection{Post hoc ROI-level Attribution Analysis}

For post hoc interpretability analysis, we use the anatomically constrained ROI-level source-pattern representations to obtain ROI-level model attributions. Specifically, we compute source-point attribution scores and aggregate them into ROI-level attribution vectors. We report both attention-based attribution and Integrated Gradients (IG) attribution (time-aggregated before ROI pooling), and form subject-level ROI-importance profiles by aggregating source-point attribution scores into ROI-importance vectors.

To quantify cross-subject consistency, we compute all pairwise Pearson correlations between subject-level ROI-importance profiles for each method, forming an $N \times N$ correlation matrix with ones on the diagonal, and report the mean off-diagonal (upper-triangular) correlation.

\begin{figure}[!t]
\centering
\begin{minipage}[t]{0.49\textwidth}
  \centering
  \includegraphics[width=\linewidth]{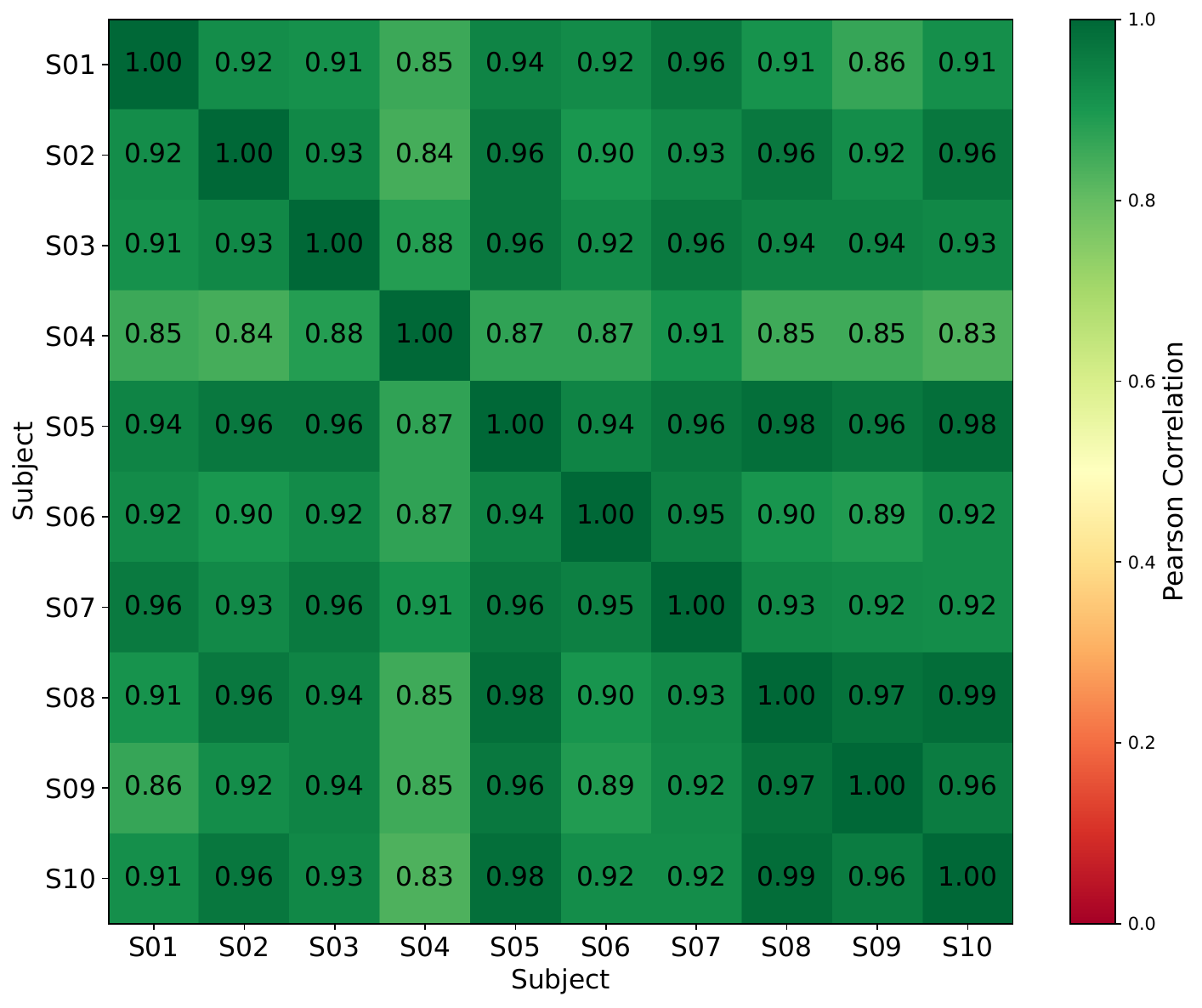}
  \par\scriptsize (a) Attention-based attribution
\end{minipage}
\hfill
\begin{minipage}[t]{0.49\textwidth}
  \centering
  \includegraphics[width=\linewidth]{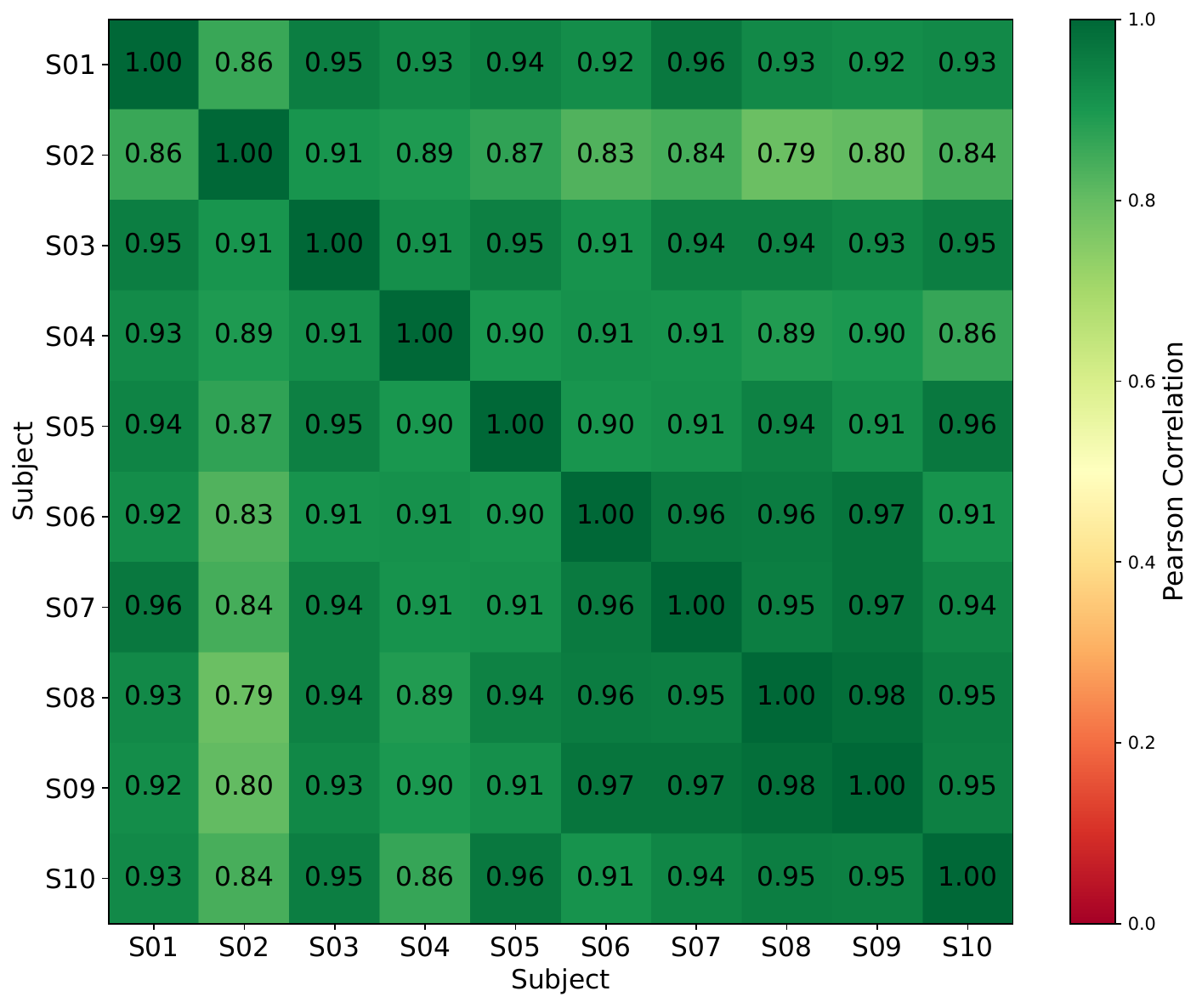}
  \par\scriptsize (b) Integrated Gradients attribution
\end{minipage}

\caption{Cross-subject consistency of ROI-importance profiles. We compute subject-level ROI-importance vectors using (a) attention-based attribution and (b) Integrated Gradients attribution, and report pairwise Pearson-correlation matrices across subjects. These profiles summarize model attributions within predefined cortical ROIs.}
\label{fig:roi_cross_subject_consistency}
\end{figure}

Despite variability in overall attribution magnitudes across subjects, the ROI-importance patterns are highly consistent. As shown in \cref{fig:roi_cross_subject_consistency}, subject-level ROI-importance vectors exhibit strong pairwise agreement for both attention-based attribution and IG attribution, with mean pairwise Pearson correlations of 0.922 and 0.915. A few subjects exhibit lower correlations with others, suggesting individual differences.

\begin{figure}[!t]
\centering
\begin{subfigure}[t]{0.49\textwidth}
  \centering
  \includegraphics[width=\linewidth]{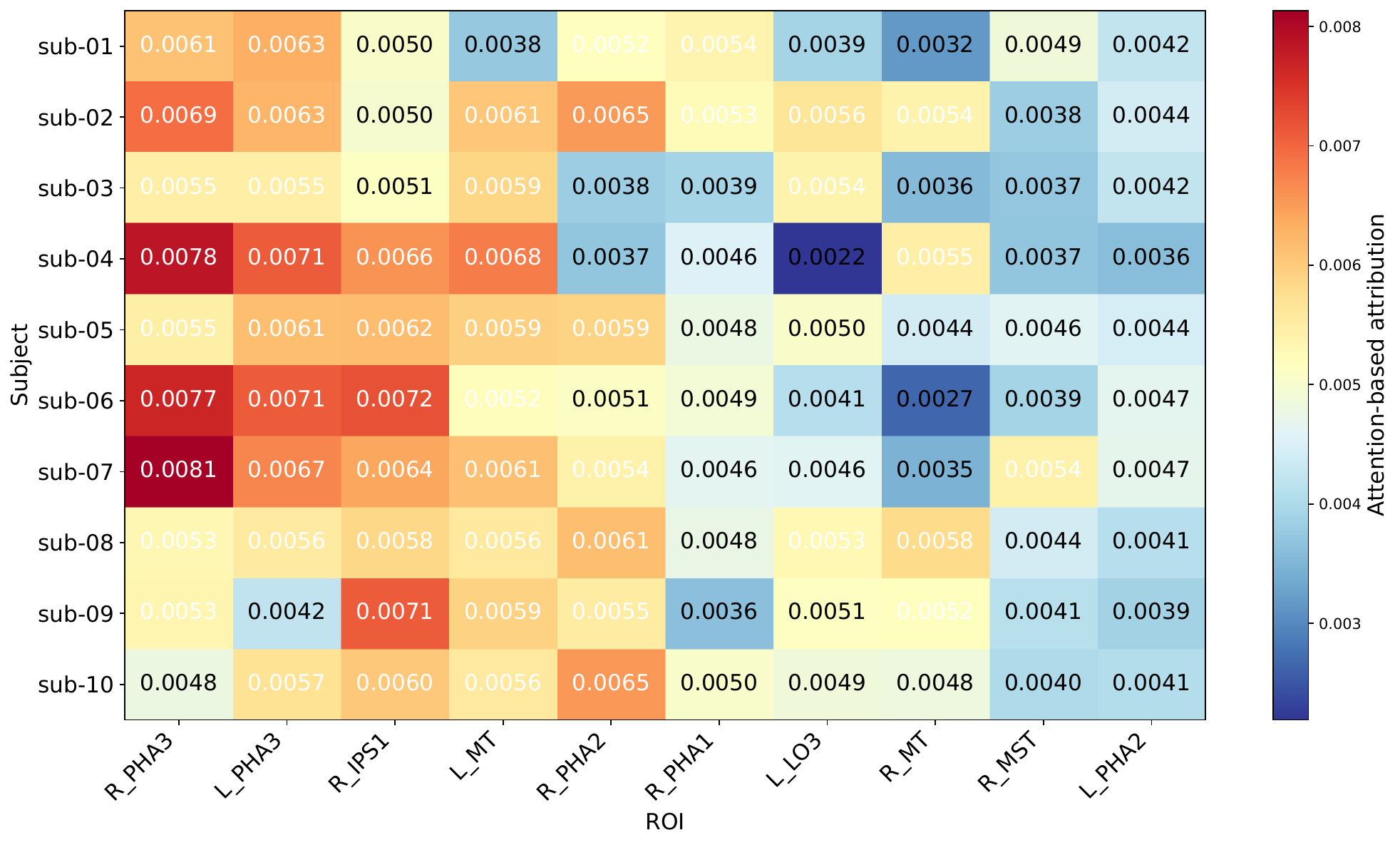}
  \caption{Attention-based attribution}
  \label{fig:attention_heatmap_subjects}
\end{subfigure}
\hfill
\begin{subfigure}[t]{0.49\textwidth}
  \centering
  \includegraphics[width=\linewidth]{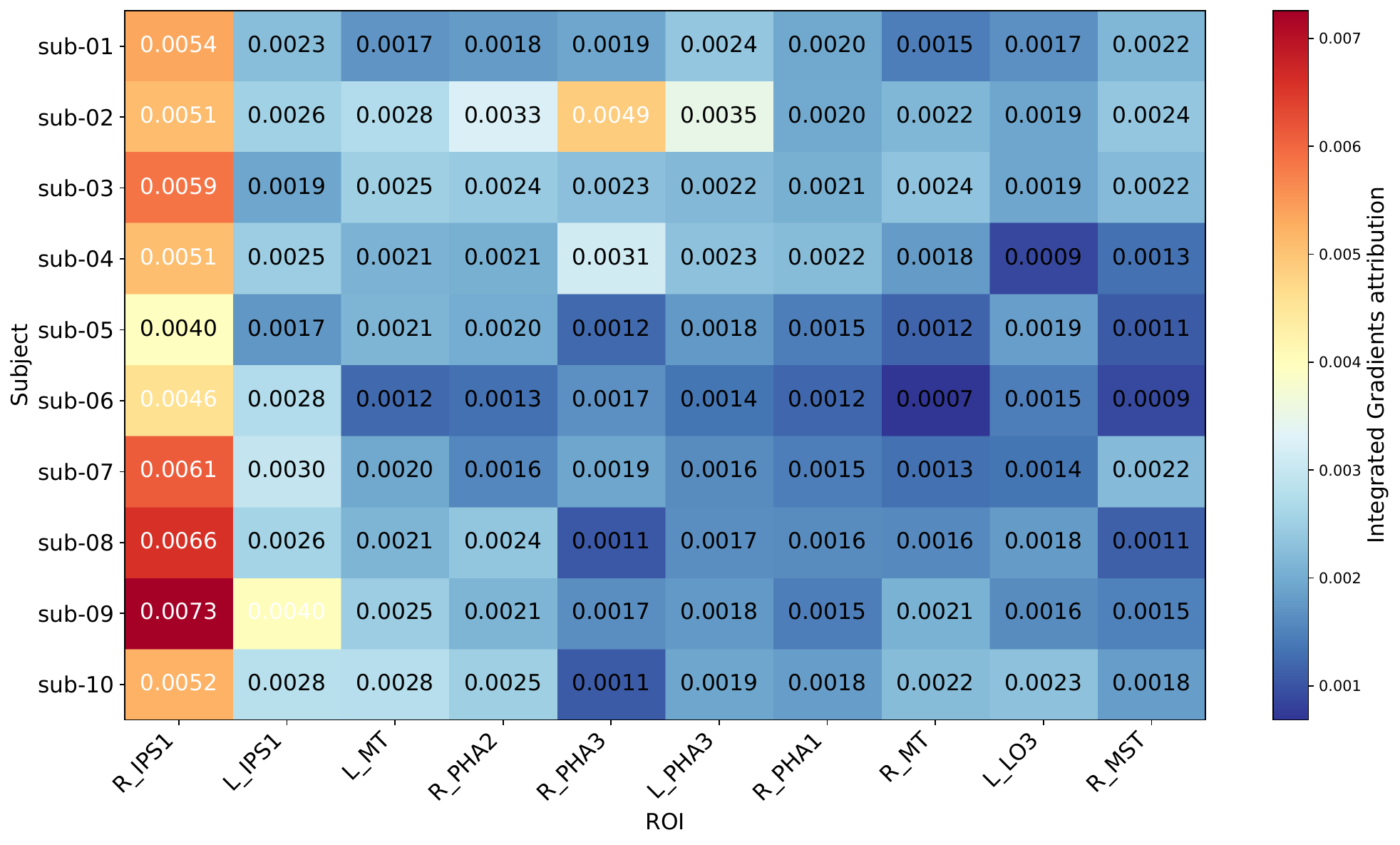}
  \caption{Integrated Gradients attribution}
  \label{fig:integrated_gradients_heatmap_subjects}
\end{subfigure}

\caption{Subject-level ROI-importance heatmaps across predefined cortical ROIs. Rows correspond to subjects and columns correspond to ROIs; values are averaged within each subject. (a) ROI importance measured by attention-based attribution. (b) ROI importance measured by Integrated Gradients attribution. Warmer colors indicate higher model attribution.}
\label{fig:roi_heatmap_subjects}
\end{figure}

The subject-level ROI heatmaps further reveal a consistent trend in which a subset of ROIs receives high model attribution (\cref{fig:roi_heatmap_subjects}). Attention-based attribution tends to emphasize higher-level visual regions in the parahippocampal cortex (PHA subregions), together with dorsal parietal IPS1 and lateral occipital areas. IG highlights a stable dominant contributor: IPS1 remains salient across subjects, while secondary contributions from MT and PHA subregions vary more noticeably, reflecting residual individual modulation. Overall, these complementary attribution patterns are consistent with model-relevant ROI-level attribution patterns over distributed higher-level visual and dorsal-stream source-pattern representations, rather than early visual ROI representations alone.

\begin{figure}[!t]
\centering

\begin{minipage}[t]{0.49\textwidth}
  \centering
  \includegraphics[width=\linewidth]{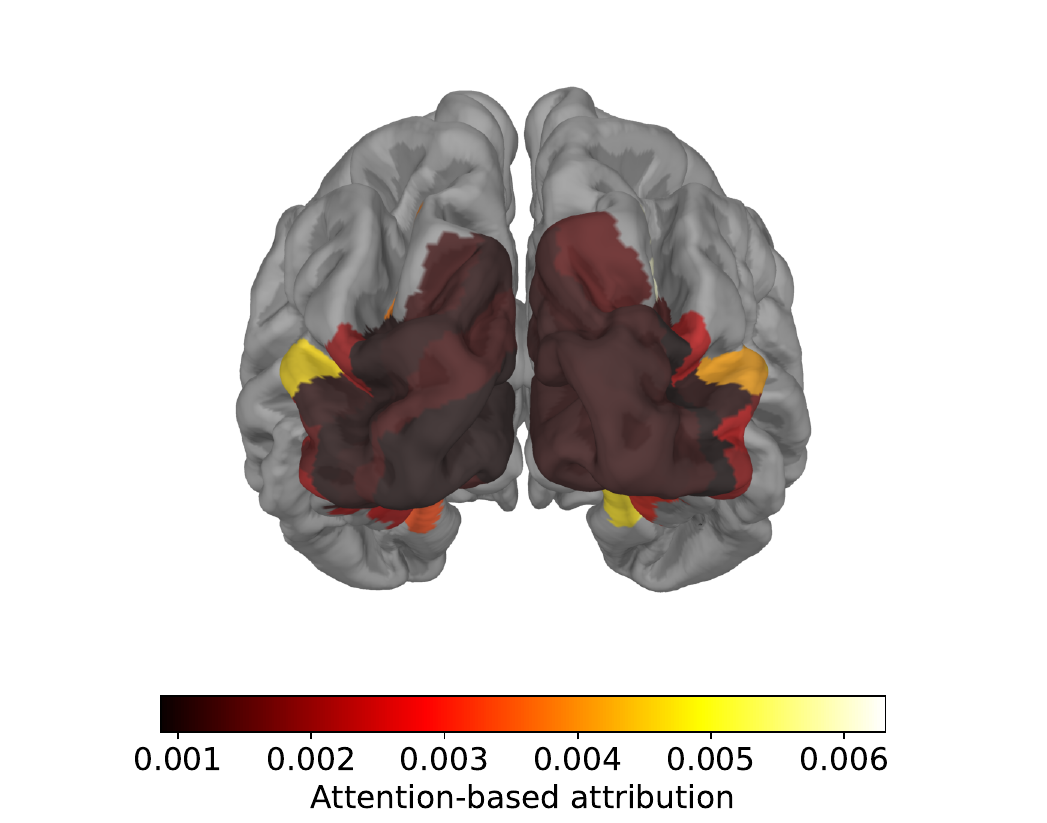}
  \par\scriptsize (a) Attention-based attribution
\end{minipage}
\hfill
\begin{minipage}[t]{0.49\textwidth}
  \centering
  \includegraphics[width=\linewidth]{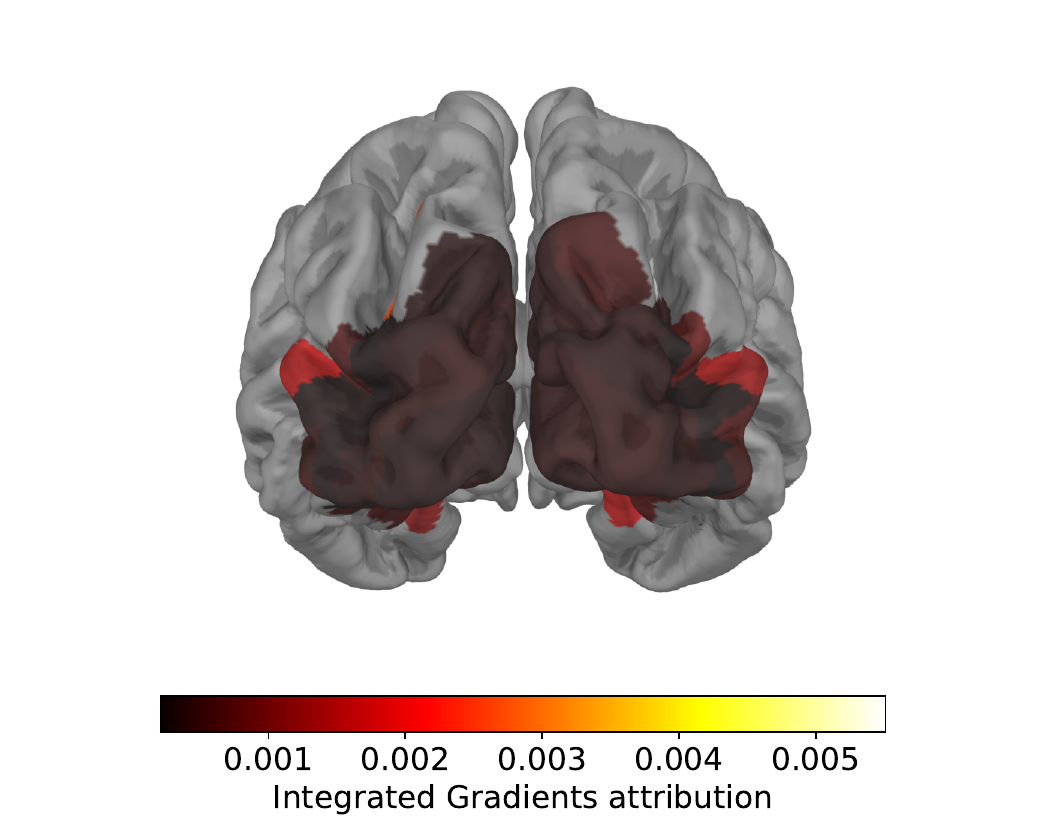}
  \par\scriptsize (b) Integrated Gradients attribution 
\end{minipage}

\caption{Cortical surface visualization of ROI-level attribution. We project (a) attention-based attribution and (b) Integrated Gradients attribution onto the cortical surface. Warmer colors indicate higher model attribution within predefined cortical ROIs.}
\label{fig:cortical_surface_vis}
\end{figure}

To summarize these attributions anatomically, we visualize ROI-level attribution on the cortical surface (\cref{fig:cortical_surface_vis}). Specifically, we map the ROI-importance heatmaps back onto their corresponding ROI locations to obtain a surface-level visualization. The surface maps highlight a compact set of visually related regions. The agreement between attention-based attribution and Integrated Gradients attribution suggests that the model consistently relies on visually relevant ROI-level source-pattern representations. These attribution patterns serve as post hoc summaries of model-relevant ROI-level source patterns.

\section{Discussion and Conclusion}

We presented an adaptive cortically constrained EEG--vision alignment method for zero-shot brain-to-image retrieval. The method combines ROI-level source-pattern representation learning with evidence-based adaptive visual supervision over detail-controlled visual targets. In this setting, the response-wise pignistic posterior is interpreted as a model-based alignment reliability estimate derived from response-wise loss and margin statistics, not as a direct neural label of visual-detail representation.

The cortically constrained source component formulates EEG--vision alignment within a structured ROI-level source-pattern representation, mitigating ambiguity from sensor-space spatial mixing and enabling post hoc summaries of model-relevant source-pattern organization. Using THINGS-EEG, the proposed approach achieved competitive within-subject zero-shot retrieval performance while producing ROI-level attribution patterns concentrated in visually relevant cortical regions. These results suggest that cortically constrained representation learning and adaptive visual supervision can jointly support retrieval-oriented EEG--vision alignment with ROI-level interpretability.

Several limitations qualify this interpretation and motivate future work. First, detail-controlled visual targets are generated by controlled blur as an operational proxy for available visual detail and should not be treated as a complete semantic hierarchy. Second, because individual structural MRIs were not available, we used template-based source reconstruction and predefined cortical ROIs; the ROI-level attribution results therefore indicate reproducible model-relevant source-pattern organization, not exact cortical generators. Third, the present analyses use repetition-averaged stimulus-specific source representations rather than raw single-presentation trials. Fourth, we did not include additional source-space control analyses such as random cortical ROI sets, non-visual cortical ROI controls, or sensor-space EEG attribution comparisons, so the ROI-level attribution findings should be interpreted descriptively. Finally, the evaluation is based on a single natural-image EEG dataset, and future work should examine broader cross-dataset validation, external measures of response-wise alignment reliability, subject-specific forward models, and extensions to more naturalistic visual stimuli.

\section*{CRediT authorship contribution statement}
\textbf{Ye Wang}: Conceptualization, Methodology, Software, Writing -- Original Draft. \textbf{Haokun Ren}: Conceptualization, Methodology, Writing -- Review \& Editing. \textbf{Wei Wu}: Data Curation, Formal Analysis. \textbf{Guoyin Wang}: Investigation, Validation. \textbf{Zhuliang Yu}: Supervision, Project Administration. \textbf{Hong Yu}: Resources, Visualization. \textbf{Ke Liu}: Supervision, Funding Acquisition.

\section*{Data Availability Statement}
The THINGS-EEG dataset used in this study is publicly available at \texttt{https://osf.io/3jk45/}. Code and pretrained models will be made available upon publication.

\section*{Ethics Statement}
The THINGS-EEG dataset was collected with informed consent from all participants and approved by the relevant institutional ethics committee. This study uses only publicly available, de-identified data.

\section*{Declaration of Competing Interests}
The authors declare that they have no known competing financial interests or personal relationships that could have appeared to influence the work reported in this paper.

\section*{Acknowledgements}
This work was partly supported by the National Natural Science Foundation of China (62136002, 62221005, 62306056, 62476034 and U24A20338).


\section*{Declaration of generative AI and AI-assisted technologies in the writing process}
During the preparation of this work the authors used OpenAI ChatGPT in order to assist with language editing, grammar refinement, and manuscript formatting. After using this tool/service, the authors reviewed and edited the content as needed and take full responsibility for the content of the publication.

\bibliographystyle{elsarticle-harv}
\bibliography{bibliography}

\appendix
\section{Experimental Details}
\subsection{Dataset Details}
\label{sec:dataset}
We performed our experiments on the THINGS-EEG dataset, a large-scale natural-image EEG corpus recorded from 10 participants during a visual recognition task. The experiment employed a Rapid Serial Visual Presentation (RSVP) paradigm with an orthogonal target-detection task to maintain attention. Each participant completed four sessions. Across sessions, the dataset includes 16,540 training image conditions, each presented four times (yielding $16{,}540 \times 4 = 66{,}160$ training trials), and 200 test image conditions, each presented 80 times (yielding $200 \times 80 = 16{,}000$ test trials). In total, the dataset contains $66{,}160 + 16{,}000 = 82{,}160$ image-evoked EEG response trials. We use the training trials for model training and the test responses only for zero-shot retrieval evaluation and post hoc posterior analyses; training and test concepts are disjoint.

For preprocessing, we used two pipelines depending on the analysis space. For the sensor-space EEG analyses, we followed the protocol of ATM-S \citep{li2024visual}. 
To prepare the data for ROI-level source analyses, we preprocessed the scalp EEG recordings before dSPM-based source reconstruction using the same steps (0.1-100~Hz bandpass filtering, retaining 63 channels at 1000~Hz, epoching 0-1000~ms with baseline correction using the 200~ms pre-stimulus interval, and downsampling to 250~Hz), but did not apply multivariate noise normalization (MVNN).
Instead, to improve the signal-to-noise ratio (SNR), we averaged repeated EEG responses to the same stimulus after dSPM-based source reconstruction.

\subsection{ROI Selection}
\label{sec:roi}
To impose cortically constrained spatial priors on EEG--vision alignment, we use cortically constrained ROI-level source modeling with a predefined set of cortical ROI labels. Table~\ref{tab:roi_labels} reports the predefined cortical ROI label set used in our experiments. ROI selection is hypothesis-driven and specified a priori, independent of downstream retrieval performance.

Following the functional-anatomical organization of the human visual system and prior vision-brain alignment work~\citep{allen2022massive,lu2023minddiffuser}, we organize the selected ROI labels into cortical ROI families according to the HCP-MMP1 parcellation to support concise reporting and interpretability. Concretely, we define the following bilateral cortical ROI families via label-name groupings: V1 (V1), V2 (V2), V3 (V3/V3A/V3B/V3CD), hV4 (V4/V4t), VO (V8/VVC/PIT), PHC (PHA1/PHA2/PHA3), MT (MT), MST (MST), LO (LO1/LO2/LO3), and IPS (IPS0-IPS5). Unless otherwise stated, we retain only the source locations whose labels belong to the predefined cortical ROI set, and discard source locations outside the predefined cortical ROI set.

By restricting modeling to this predefined cortical ROI label set (rather than using unconstrained sensor-space EEG recordings), we narrow the spatial hypothesis space of EEG--vision alignment and reduce ambiguity induced by volume-conduction-related spatial mixing.

\begin{table}[htbp]
\centering
\caption{Predefined cortical ROI families and corresponding HCP-MMP1 label names.}
\label{tab:roi_labels}
\begin{tabular}{ll}
\toprule
Cortical ROI family & HCP-MMP1 label names \\
\midrule
V1  & V1 \\
V2  & V2 \\
V3  & V3, V3A, V3B, V3CD \\
hV4 & V4, V4t \\
VO  & V8, VVC, PIT \\
PHC & PHA1, PHA2, PHA3 \\
MT  & MT \\
MST & MST \\
LO  & LO1, LO2, LO3 \\
IPS & IPS0, IPS1, IPS2, IPS3, IPS4, IPS5 \\
\bottomrule
\end{tabular}
\end{table}

\subsection{Implementation Details and Hyperparameters}

\label{sec:implementation}

\textbf{Environment.} Our implementation is based on Python 3.10.16, CUDA 12.2, and PyTorch 2.7.0. Experiments are conducted on a server equipped with an AMD EPYC 7742 64-Core Processor and 256 GB of system memory, using two NVIDIA L40 GPUs (48GB).

\textbf{Training Configuration.} We perform five independent training runs and report the average. Unless otherwise stated, we adopt a consistent training protocol, while adjusting the batch size and number of training epochs for different evaluation settings. For the intra-subject setting, we train all models for 200 epochs with a batch size of 64. For the inter-subject setting, we train all models for 50 epochs with a batch size of 128. In both settings, we use the AdamW optimizer with a learning rate of 1e-4.

\textbf{Hyperparameters.} For the detail-controlled visual targets (Sec.~\ref{sec:mrvt}), we follow the uncertainty-aware blur prior of UBP \citep{wu2025bridging} and keep its parameterization unchanged (e.g., foveation and decay settings). The only difference is that the low-/medium-/high-detail visual targets are generated by different blur-kernel radii, $\{r_L,r_M,r_H\}=\{61,21,3\}$, with $r_H=3$ being closest to the sharp image. For each response-wise alignment statistic $x\in\{\ell,\mu\}$ (Sec.~\ref{sec:elra}), we track EMA-smoothed running quantiles $\theta_x^{(30)}$ and $\theta_x^{(70)}$ (0.3/0.7) with momentum $\alpha$, and normalize to $[0,1]$. Table~\ref{tab:hparams} lists all hyperparameters for response-wise quantile calibration, conflict-to-ignorance evidence conversion, and evidence discounting.


\begin{table}[t]
\centering
\scriptsize
\setlength{\tabcolsep}{3.5pt}
\renewcommand{\arraystretch}{1.05}
\caption{Hyperparameters used in the proposed adaptive cortically constrained EEG--vision alignment method.}
\label{tab:hparams}
\begin{tabularx}{\linewidth}{
  >{\raggedright\arraybackslash}p{2.2cm}
  >{\raggedright\arraybackslash}p{2.3cm}
  c
  c
  >{\raggedright\arraybackslash}X
}
\toprule
\textbf{Component} & \makecell[c]{\textbf{Hyper-}\\\textbf{parameter}} & \textbf{Symbol} & \textbf{Value} & \textbf{Description} \\
\midrule
\makecell[l]{Response-wise\\Quantile\\Calibration}
& Quantile low & $q_{\rm low}$ & 0.3
& Lower running quantile for $\theta_x^{(30)}$, $x\in\{\ell,\mu\}$. \\

\makecell[l]{Response-wise\\Quantile\\Calibration}
& Quantile high & $q_{\rm high}$ & 0.7
& Upper running quantile for $\theta_x^{(70)}$, $x\in\{\ell,\mu\}$. \\

\makecell[l]{Response-wise\\Quantile\\Calibration}
& EMA momentum & $\alpha$ & 0.95
& Momentum for updating running quantiles. \\

\midrule
Alignment-evidence Mapping
& Temperature & $T$ & 0.5
& Controls the sharpness of alignment-evidence mapping. \\

\midrule
\makecell[l]{Conflict-to-\\Ignorance\\Conversion}
& Base offset & $\beta_0$ & 0.15
& Base term in the ignorance mass assignment. \\

\makecell[l]{Conflict-to-\\Ignorance\\Conversion}
& Inconsistency weight & $\beta_1$ & 0.3
& Weight on inconsistency when assigning ignorance mass. \\

\makecell[l]{Conflict-to-\\Ignorance\\Conversion}
& Ignorance floor & $\epsilon$ & 0.05
& Lower bound for numerical stability. \\

\midrule
Evidence Discounting
& Evidence discount factor & $\delta$ & 0.95
& Epoch-wise evidence discounting to prevent stale alignment evidence from dominating. \\
\bottomrule
\end{tabularx}
\end{table}

\section{Additional Results}

\subsection{Effect of Visual Encoders}
\label{sec:visual_encoders}
We evaluate the performance of the proposed adaptive cortically constrained EEG--vision alignment method under different frozen visual encoders. Specifically, we use visual embeddings extracted by DINOv2 ViT-B/14, DINO ViT-B/8, DINO ViT-B/16, and DINO RN50, while keeping all other training settings unchanged. Table~\ref{tab:backbone} reports the intra-subject averaged Top-1 and Top-5 zero-shot retrieval accuracy. The proposed approach achieves the best results with DINO RN50 56.7\%/83.0\% (Top-1/Top-5).

\begin{table}[h!]
\centering
\small
\setlength{\tabcolsep}{6pt}
\renewcommand{\arraystretch}{1.15}
\caption{Effect of visual encoders in the proposed adaptive cortically constrained EEG--vision alignment method. Results are reported as intra-subject averaged Top-1 and Top-5 zero-shot retrieval accuracy.}
\label{tab:backbone}
\begin{tabular}{lcc}
\toprule
\textbf{Visual Encoder} & \multicolumn{2}{c}{\textbf{Avg.\ Acc.\ (\%)}} \\
\cmidrule(lr){2-3}
 & \textbf{Top-1} & \textbf{Top-5} \\
\midrule
DINOv2 ViT-B/14 & 23.7 & 46.5 \\
DINO ViT-B/8   & 51.6 & 79.4 \\
DINO ViT-B/16  & 53.5 & 81.2 \\
DINO RN50      & \textbf{56.7} & \textbf{83.0} \\
\bottomrule
\end{tabular}
\end{table}

\subsection{Comparison of EEG Encoders}

\label{sec:brain_encoders}
We compare our proposed Neuro-ROI Attention Encoder with representative EEG encoders while keeping all
other components and training settings unchanged. As shown in Table~\ref{tab:brain-compare}, Neuro-ROI Attention Encoder achieves the best overall performance.


\begin{table}[htbp]
\caption{Comparison with EEG encoders on THINGS-EEG 200-way zero-shot retrieval. We report average intra-subject Top-1/Top-5 accuracy.}
\centering
\small
\begin{tabular}{lcc}
\toprule
\textbf{EEG Encoders} & \multicolumn{2}{c}{\textbf{Avg. Acc. (\%)}} \\
\cmidrule(lr){2-3}
 & \textbf{Top-1} & \textbf{Top-5} \\
\midrule
DeepNet      & 35.8 & 65.6 \\
EEGProject   & 46.4 & 75.8 \\
ShallowNet   & 47.2 & 76.5 \\
EEGNet       & 51.3 & 78.6 \\
TSConv       & 51.5 & 79.8 \\
\textbf{Neuro-ROI Attention Encoder} & \textbf{56.7} & \textbf{83.0} \\
\bottomrule
\end{tabular}
\label{tab:brain-compare}
\end{table}

\end{document}